\documentclass[10pt]{article}

\usepackage[utf8]{inputenc}
\usepackage[T1]{fontenc}
\usepackage{graphicx}
\usepackage{amsmath, amssymb}
\usepackage{booktabs}
\usepackage{caption}
\usepackage{subcaption}
\usepackage{url}
\usepackage{hyperref}
\usepackage{authblk}
\usepackage[font=small,labelfont=bf]{caption}
\usepackage[margin=1in]{geometry}
\usepackage{algorithm}
\usepackage{algpseudocode}
\usepackage{listings}
\usepackage{xcolor}

\definecolor{codegreen}{rgb}{0,0.6,0}
\definecolor{codegray}{rgb}{0.5,0.5,0.5}
\definecolor{codepurple}{rgb}{0.58,0,0.82}
\definecolor{backcolour}{rgb}{0.95,0.95,0.92}

\lstdefinestyle{mystyle}{
    backgroundcolor=\color{backcolour},   
    commentstyle=\color{codegreen},
    keywordstyle=\color{magenta},
    numberstyle=\tiny\color{codegray},
    stringstyle=\color{codepurple},
    basicstyle=\ttfamily\footnotesize,
    breakatwhitespace=false,         
    breaklines=true,                 
    captionpos=b,                    
    keepspaces=true,                 
    numbers=left,                    
    numbersep=5pt,                  
    showspaces=false,                
    showstringspaces=false,
    showtabs=false,                  
    tabsize=2
}
\begin{document}

\title{ZACH-ViT: A Zero-Token Vision Transformer with ShuffleStrides Data Augmentation for Robust Lung Ultrasound Classification}

\author[1,2,*]{Athanasios Angelakis}
\author[3]{Amne Mousa}
\author[3]{Micah L. A. Heldeweg}
\author[3]{Laurens A. Biesheuvel}
\author[3]{Mark A. Haaksma}
\author[3]{Jasper M. Smit}
\author[3]{Pieter R. Tuinman}
\author[3]{Paul W. G. Elbers}

\affil[1]{Department of Epidemiology and Data Science, Amsterdam UMC, University of Amsterdam, The Netherlands}
\affil[2]{Data Science Center, University of Amsterdam, The Netherlands}
\affil[3]{Department of Intensive Care Medicine, Amsterdam UMC, University of Amsterdam \& Vrije Universiteit Amsterdam, The Netherlands}

\affil[*]{\textit{Corresponding author:} a.angelakis@amsterdamumc.nl\\
ath.angelakis@gmail.com\\ ORCID: 0000-0003-1226-9560
}

\date{}

\maketitle

\begin{abstract}
Differentiating cardiogenic pulmonary oedema (CPE) from non-cardiogenic and structurally normal lungs in lung ultrasound (LUS) videos remains a major challenge in critical care imaging. The non-cardiogenic group (Class 0) in this study encompassed a heterogeneous mixture of non-cardiogenic inflammatory pathology (NCIP/ARDS-like), interstitial lung disease (ILD), and healthy lung patterns, reflecting the broad visual variability encountered in clinical practice. This heterogeneity substantially increases the difficulty of automated classification, as these entities share overlapping B-line and pleural artefacts on ultrasound.

We introduce \textbf{ZACH-ViT} (\textbf{Z}ero-token \textbf{A}daptive \textbf{C}ompact \textbf{H}ierarchical Vision Transformer), a Vision Transformer variant that eliminates both positional embeddings and the [CLS] token, rendering it fully permutation-invariant and suitable for unordered medical image data. To enhance generalization, we propose \textbf{ShuffleStrides Data Augmentation (SSDA)}, a structured augmentation strategy that permutes probe-view sequences and frame orders while preserving anatomical validity.

From a computer vision perspective, this setting presents a challenging testbed for model robustness under extreme intra-class heterogeneity and domain shift—conditions increasingly relevant in real-world deployment. Unlike curated benchmarks (e.g., ImageNet), clinical ultrasound exhibits non-stationary spatial layouts, variable probe geometries, and overlapping visual patterns across diagnostic categories.

We evaluate ZACH-ViT against nine state-of-the-art baselines on 380 LUS videos from 95 critically ill patients. Despite the increased heterogeneity of Class 0, ZACH-ViT achieves the highest validation and test ROC-AUC (0.80 and 0.79) with balanced sensitivity (0.60) and specificity (0.91), while all competing models collapse to trivial classification. This demonstrates that architectural alignment with data structure is more critical than model scale in small-data medical settings. ZACH-ViT trains \textbf{1.35× faster} than Minimal ViT with \textbf{2.5× fewer parameters}, making it suitable for real-time clinical deployment. These results demonstrate that architectural parsimony aligned with data structure can outperform complex models in medical imaging. Code and package: \url{https://github.com/Bluesman79/ZACH-ViT} , \texttt{pip install zachvit}.  
\end{abstract}


\section{Introduction}

Pulmonary oedema is one of the leading causes of respiratory failure among critically ill patients~\cite{ware2005acute,bellani2016epidemiology}. The ability to differentiate \emph{cardiogenic} from \emph{non-cardiogenic} (e.g., ARDS-related) pulmonary oedema is crucial because therapeutic management fundamentally differs between conditions driven by cardiac dysfunction and those caused by increased vascular permeability~\cite{meyer2021ards,smit2023lung}. 

Lung ultrasound (LUS) has become a standard bedside imaging tool in intensive care due to its portability, safety, and capability for real-time visualization~\cite{lichtenstein2014ten,heldeweg2023impact}. However, despite its widespread adoption, LUS interpretation remains a qualitative, operator-dependent process, with poor reproducibility across centers and examiners~\cite{heldeweg2022lung,heldeweg2023gestalt}. These limitations make the development of automated, data-driven interpretation systems both clinically necessary and methodologically challenging.

In real-world clinical practice, the non-cardiogenic (Class 0) category is heterogeneous, encompassing multiple pulmonary states beyond acute inflammatory or permeability-related oedema. Specifically, Class 0 in this work includes cases of \emph{non-cardiogenic inflammatory pathology} (NCIP, ARDS-like), \emph{interstitial lung disease} (ILD), and \emph{structurally normal or healthy} lungs. This spectrum reflects the full diagnostic landscape encountered in bedside LUS~\cite{volpicelli2020international,soldati2020standardization,mongodi2025esicm}. Such heterogeneity makes the classification task more difficult: the model must learn to distinguish cardiogenic oedema not only from ARDS-like pathology but also from interstitial fibrotic patterns and normal aerated lungs, each with distinct and overlapping sonographic signatures. From a modeling perspective, this setting provides a stronger and more realistic test of robustness than the simpler binary CPE–NCIP separation.

While motivated by lung ultrasound, the design principles of ZACH-ViT, permutation invariance, absence of positional bias, and global pooling over local features, are broadly applicable to any vision task involving unordered or weakly ordered image collections. Examples include multi-view satellite imagery, bag-of-patches histopathology, robotic tactile sensing arrays, or any setting where spatial arrangement is non-diagnostic or highly variable.

\subsection{Deep Learning and the Challenge of Domain Shift}

Recent advances in deep learning have produced impressive results across radiology and pathology~\cite{campanella2019clinical,esteva2017dermatologist}. Convolutional neural networks (CNNs) and Vision Transformers (ViTs)~\cite{dosovitskiy2020image,liu2021swin,liu2022convnet} have achieved high accuracy on large, homogeneous datasets such as ImageNet or histopathology slides. However, their performance deteriorates on small, irregular, and domain-shifted datasets, which are common in medical imaging. 

LUS data are particularly challenging: image appearance varies with probe type, position, patient anatomy, and respiratory motion. These sources of variability introduce non-stationary spatial and temporal patterns that violate the assumptions of both convolutional locality and transformer positional regularity. As a result, standard deep models either overfit to superficial features or collapse to trivial outputs under small-sample conditions.

\subsection{Proposed Framework}

To address these issues, we propose \textbf{ZACH-ViT}, the \textbf{Z}ero-token, \textbf{A}daptive, \textbf{C}ompact, and \textbf{H}ierarchical Vision Transformer. ZACH-ViT is explicitly designed for small, heterogeneous medical imaging datasets, emphasizing model stability and generalization over scale. The architecture eliminates positional embeddings and class tokens, two components that can inadvertently introduce spatial bias in unordered medical data, and replaces them with adaptive residual projections and global pooling mechanisms that preserve permutation invariance. We refer to this design as “zero-token” because it contains no learnable class token and no positional encodings, relying solely on patch embeddings and global pooling for prediction.

To further strengthen robustness, we introduce \textbf{ShuffleStrides Data Augmentation (SSDA)}, a structured augmentation framework tailored to the spatiotemporal nature of LUS data. SSDA creates clinically valid diversity by permuting transducer-view sequences and frame orderings, ensuring that augmented data remain anatomically meaningful while improving statistical generalization. 

Both the ZACH-ViT architecture and the ShuffleStrides Data Augmentation (SSDA) framework were developed by the first author and are available as a Python package (\texttt{pip install zachvit})~\cite{angelakis2025zachvit} and open-source implementation at \href{https://github.com/Bluesman79/ZACH-ViT}{https://github.com/Bluesman79/ZACH-ViT}.

\subsection{Study Objectives and Contributions}

We evaluate ZACH-ViT on 380 lung ultrasound videos from 95 critically ill patients. Using semi-supervised augmentation regimes (0-SSDA and 0\_2-SSDA), we benchmark our model against nine state-of-the-art baselines, including CNN, Transformer, and Multiple-Instance Learning (MIL) architectures. Only ZACH-ViT achieves stable convergence and balanced performance across validation and test sets.

Our main contributions are as follows:
\begin{itemize}
    \item We propose \textbf{ZACH-ViT}, a minimal yet powerful Vision Transformer architecture that eliminates positional embeddings and class tokens, enabling robust learning from unordered medical images.
    \item We introduce \textbf{ShuffleStrides Data Augmentation (SSDA)}, a structured augmentation method that maintains anatomical and temporal consistency in ultrasound data.
    \item We explicitly address the realistic classification problem of distinguishing cardiogenic oedema from a composite non-cardiogenic class that includes NCIP, ILD, and healthy lungs, offering a benchmark more representative of clinical variability.
    \item We release the complete open-source implementation of \textbf{ZACH-ViT} and its accompanying \textbf{ShuffleStrides Data Augmentation (SSDA)} framework, including preprocessing and training pipelines, to foster reproducibility and enable adaptation to other domains involving unordered or weakly ordered image data.
\end{itemize}

\section{Methods}

\subsection{Dataset and Study Design}

We conducted a retrospective study using lung ultrasound (LUS) data from 95 critically ill patients admitted to Amsterdam University Medical Centers between 2016 and 2020. The study was approved by the institutional ethics committee (reference: 2021.0102). Each patient contributed up to four lung ultrasound videos corresponding to anterolateral thoracic regions, resulting in a total of 380 videos. Recordings were obtained using a FUJIFILM SonoSite Edge II device operating at 30\,frames per second with a convex probe.

Inclusion criteria required that patients had a confirmed diagnosis of either \emph{cardiogenic pulmonary oedema} (CPE) or \emph{non-cardiogenic pathology}. In contrast to earlier formulations of this task, the non-cardiogenic class (hereafter \textbf{Class 0}) was deliberately defined to reflect the heterogeneity of real-world bedside imaging. Specifically, Class 0 included patients with \emph{non-cardiogenic inflammatory pathology} (NCIP or ARDS-like patterns), \emph{interstitial lung disease} (ILD), and \emph{lungs without structural abnormality} (healthy). This spectrum captures the range of acoustic and structural variability encountered in clinical practice, making the task more realistic and diagnostically demanding~\cite{volpicelli2020international,soldati2020standardization,mongodi2025esicm}.  
From a modeling perspective, this expanded Class 0 composition forces the classifier to discriminate cardiogenic oedema not merely from other oedematous states but from multiple physiological regimes that share overlapping sonographic signatures.

All videos were fully de-identified and stored in DICOM format. Each video was subsequently decomposed into grayscale frame sequences for further processing.

To ensure reproducibility and patient-level independence, the dataset was divided into three disjoint subsets:
61 patients for training, 18 for validation, and 16 for testing. The splits maintained a balanced representation of cardiogenic cases in each subset (5 per evaluation set). Baseline demographic and clinical characteristics are reported in Table~\ref{tab:demographics}.

\subsection{Preprocessing and Region of Interest (ROI)}
Each ultrasound video underwent standardized preprocessing prior to model input. From each frame, a
region of interest (ROI) centered on the pleural line was extracted based on expert-defined coordinates. This
step ensured consistent focus on diagnostically relevant anatomical structures across patients and probe
positions, in line with principles from ultrasound elasticity mapping \cite{bercoff2004supersonic,sigrist2017ultrasound}. 

The preprocessing pipeline consisted of the following operations:
\begin{enumerate}
    \item Conversion to grayscale and intensity normalization to the [0,1] range.
    \item Rescaling each frame to 224 × 224 pixels to match transformer patch dimensions.
    \item Zeroing all pixels with intensity values below 93 (on the 0–255 scale) to suppress low-frequency background noise.
    \item Cropping to the pleural ROI and retaining high-frequency components such as B-lines, pleural thickening, and subpleural consolidations.
\end{enumerate}

\begin{figure}[htbp]
\centering
\includegraphics[width=0.9\linewidth]{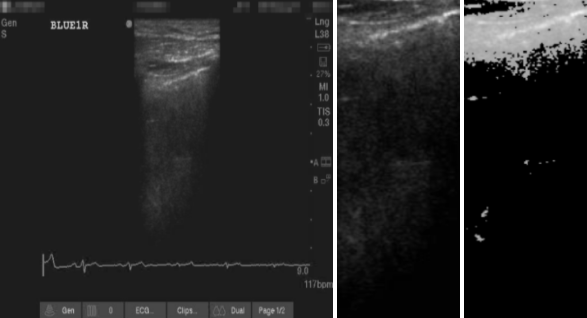}
\caption{Region of interest (ROI) extraction pipeline. (Left) Original LUS frame; (Middle) intensity thresholding to suppress background; (Right) pleural-line-centered crop retaining B-lines and pleural artifacts. This standardized ROI ensures focus on diagnostically relevant structures across patients.}
\label{fig:roi}
\end{figure}

\begin{figure}[htbp]
\centering
\includegraphics[width=0.9\linewidth]{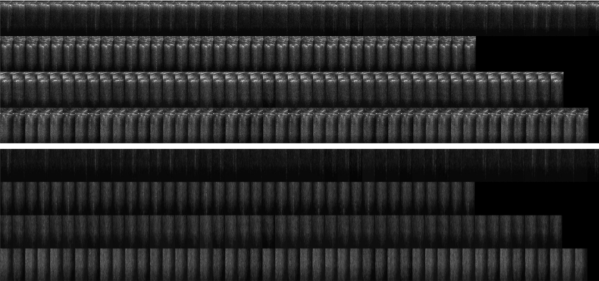}
\caption{Height Reduction to 50\% and final ROI.}
\label{fig:hightcrop}
\end{figure}

This preprocessing emphasized the key ultrasound patterns used for differential diagnosis while minimizing irrelevant artifacts such as motion blur and rib shadowing. The approach is consistent with elasticity-informed ROI extraction, ensuring the anatomical and physical integrity of the acoustic signal \cite{bercoff2004supersonic,sigrist2017ultrasound}. An example ROI extraction pipeline is shown in Figures~\ref{fig:roi} and~\ref{fig:hightcrop}.

\subsection{Video-to-Image Transformation}

Because the number of frames per LUS video and the probe position differ between patients, we implemented a deterministic set of video-to-image transformations that convert entire video sequences into 2D tensor representations. These transformations standardize temporal and spatial variability while maintaining clinical interpretability.

\begin{figure}[htbp]
\centering
\includegraphics[width=0.9\linewidth]{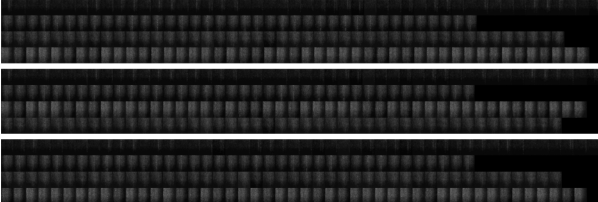}
\caption{Video-to-image transformation modes. VI: horizontal frame concatenation; SVI: shuffled frames; VIS: vertical stacking of four anatomical views. ShuffleStrides Data Augmentation (SSDA) permutes the order of these views (e.g., [1,2,3,4] → [1,2,4,3]) while preserving anatomical validity.}
\label{fig:vis}
\end{figure}

Five transformation modes were defined:

\begin{itemize}
    \item \textbf{VI (Video-Image):} Horizontal concatenation of all frames from a single video to produce a composite image.
    \item \textbf{SVI (Shuffled Video-Image):} Randomized frame order applied to VI representations; serves as data augmentation while preserving frame-level content.
    \item \textbf{VIS (Video-Image Strides):} Concatenation of four probe-view videos (one per thoracic region) into a vertically stacked \emph{stride image} (see Figure~\ref{fig:vis}). Each stride corresponds to a standardized probe placement.
    \item \textbf{0-SSDA:} The \textbf{ShuffleStrides Data Augmentation (SSDA)} baseline, consisting of 24 possible permutations of probe-view order. Each permutation simulates clinically valid yet distinct probe scanning sequences.
    \item \textbf{0\_2-SSDA:} An extension of 0-SSDA that introduces intra-view frame shuffling using ten prime-number random seeds (2, 3, 5, 7, 11, 13, 17, 19, 23, 29), producing up to twelve-fold data expansion.
\end{itemize}

Each transformation mode preserves local spatial structures and diagnostic ultrasound patterns (e.g., comet-tail artifacts, pleural irregularities) while maximizing intra-class variability. The resulting 2D images form the standardized inputs for both transformer and convolutional networks.

To mitigate class-dependent bias and improve generalization, the ShuffleStrides Data Augmentation (SSDA) framework draws inspiration from recent work highlighting that augmentation and regularization effects can vary by class \cite{balestriero2022effects,angelakis2024data}. By constraining permutations to clinically valid probe-view sequences, SSDA ensures that each augmented sample remains anatomically interpretable while enhancing representation diversity.

\begin{algorithm}[htbp]
\caption{ShuffleStrides Data Augmentation (SSDA)}
\label{alg:ssda}
\begin{algorithmic}[1]
\Require Set of videos $\mathcal{V} = \{V_1, V_2, V_3, V_4\}$ for four transducer positions
\Ensure Augmented dataset $\mathcal{D}_{\text{aug}}$
\State $\mathcal{D}_{\text{aug}} \leftarrow \emptyset$
\For{$\pi \in \text{Permutations}([1,2,3,4])$} \Comment{24 possible permutations}
    \State $\mathbf{X}_{\text{VIS}} \leftarrow \text{Concat}_{\text{vertical}}(\text{Resize}(V_{\pi(1)}), \dots, \text{Resize}(V_{\pi(4)}))$
    \State $\mathcal{D}_{\text{aug}} \leftarrow \mathcal{D}_{\text{aug}} \cup \{\mathbf{X}_{\text{VIS}}\}$
\EndFor
\For{$s \in \{2, 3, 5, 7, 11, 13, 17, 19, 23, 29\}$} \Comment{Prime-number seeds}
    \For{each video $V_i$}
        \State $V_i' \leftarrow \text{ShuffleFrames}(V_i, \text{seed}=s)$
    \EndFor
    \State Repeat permutation step with shuffled videos
\EndFor
\State \Return $\mathcal{D}_{\text{aug}}$
\end{algorithmic}
\end{algorithm}

Examples of the transformation outputs are shown in Figure~\ref{fig:vis}.

\subsection{The ZACH-ViT Architecture}
\label{subsec:zachvit}

We introduce \textbf{ZACH-ViT} (\textbf{Z}ero-token \textbf{A}daptive \textbf{C}ompact \textbf{H}ierarchical Vision Transformer), a new transformer architecture specifically optimized for learning from unordered and small-scale medical imaging datasets. ZACH-ViT redefines the standard Vision Transformer (ViT) architecture through three key innovations that collectively improve robustness and training stability:

\paragraph{1. Complete Elimination of Positional Bias.}
Conventional ViTs encode spatial order through positional embeddings:
\[
\mathbf{z}_0 = [\mathbf{x}_{\text{cls}}; \mathbf{x}_p^1\mathbf{E}; \cdots; \mathbf{x}_p^N\mathbf{E}] + \mathbf{E}_{\text{pos}}.
\]
However, for ultrasound images, spatial order is neither consistent nor diagnostic. ZACH-ViT removes $\mathbf{E}_{\text{pos}}$ entirely, producing a model that is inherently permutation-invariant and ideally suited for ShuffleStrides-based augmentations.

\paragraph{2. Dynamic Adaptive Residual Connections.}
Traditional transformers assume fixed feature dimensions across layers, which can destabilize training when applied to variable feature spaces. ZACH-ViT employs adaptive residuals that automatically project features to matching dimensions:
\begin{lstlisting}[language=Python]
y = LayerNorm(x)
y = MultiHeadAttention(y, y)
y = Dropout(0.1)(y)
if x.shape[-1] != y.shape[-1]:
    x = Dense(y.shape[-1])(x)  # Dynamic projection layer
x = x + y  # Residual connection
\end{lstlisting}
This mechanism ensures gradient stability even under irregular data distributions.

\paragraph{3. Global Pooling Instead of a Class Token.}
Rather than introducing a learnable [CLS] token, we compute a global average representation:
\[
\mathbf{h} = \text{GlobalAveragePooling1D()}(\mathbf{x}_L) = \frac{1}{N}\sum_{i=1}^{N} \mathbf{x}_L^{(i)}.
\]
This simplification reduces overfitting risk and improves interpretability without compromising discriminative power.

\subsection{Architectural Superiority Over Standard ViTs}
\label{subsec:superiority}

ZACH-ViT introduces mathematically grounded simplifications that enhance both efficiency and generalization.

\paragraph{Mathematical Formulation.}
Given input $\mathbf{X} \in \mathbb{R}^{224 \times 224 \times 3}$, the patch extraction step yields:
\[
\mathbf{P} = \text{Reshape}(\text{extract\_patches}(\mathbf{X})) \in \mathbb{R}^{196 \times 768},
\]
followed by dimensionality reduction:
\[
\mathbf{Z}_0 = \text{Dense}(128)(\mathbf{P}),
\]
without adding any positional embeddings.

\paragraph{Progressive Feature Refinement.}
Each transformer block $l$ operates as:
\begin{align*}
\mathbf{Z}_l' &= \text{LayerNorm}(\mathbf{Z}_{l-1}), \\
\mathbf{A}_l &= \text{MultiHeadAttention}(\mathbf{Z}_l', \mathbf{Z}_l'), \\
\mathbf{Z}_l^{\text{mid}} &= \text{AdaptiveAdd}(\mathbf{Z}_{l-1}, \mathbf{A}_l), \\
\mathbf{Z}_l'' &= \text{LayerNorm}(\mathbf{Z}_l^{\text{mid}}), \\
\mathbf{F}_l &= \text{Dense}(\text{units}_l)(\mathbf{Z}_l''), \\
\mathbf{Z}_l &= \text{AdaptiveAdd}(\mathbf{Z}_l^{\text{mid}}, \mathbf{F}_l),
\end{align*}
where $\text{AdaptiveAdd}$ implements the dimension-matching residual connection.

\paragraph{Parameter Efficiency.}
ZACH-ViT contains 0.25\,M parameters compared with 0.62\,M in the Minimal ViT configuration (a minimal ViT), representing a 60\% reduction in complexity while achieving a higher validation ROC-AUC (0.80 vs.\ 0.58). Our Minimal ViT implements a streamlined Vision Transformer with 8 layers, 64-dim embeddings, and positional encodings, but omits the [CLS] token—serving as a strong yet lightweight baseline for small-scale medical imaging.
This highlights that architectural parsimony, when aligned with the data domain, can yield superior performance in small-scale medical imaging tasks.

\subsection{Theoretical Motivation}
\label{subsec:theory}

The success of ZACH-ViT can be understood through the lens of permutation invariance in medical ultrasound. Let $\mathcal{T}$ be the set of all clinically valid permutations of transducer positions. For any permutation $\pi \in \mathcal{T}$, the diagnostic content remains invariant:
\[
P(y|\mathbf{X}) = P(y|\pi(\mathbf{X})) \quad \forall \pi \in \mathcal{T}
\]
Standard ViTs violate this invariance through positional embeddings $\mathbf{E}_{\text{pos}}$, whereas ZACH-ViT's positional-embedding-free design naturally satisfies:
\[
f_{\text{ZACH-ViT}}(\mathbf{X}) = f_{\text{ZACH-ViT}}(\pi(\mathbf{X}))
\]
This architectural alignment with domain symmetry explains ZACH-ViT's superior generalization on ultrasound data where probe positioning varies substantially across operators.

\subsection{Baselines and Comparative Models}

To evaluate performance comprehensively, we benchmarked ZACH-ViT against nine contemporary architectures across three families:
\begin{itemize}
\item \textbf{CNN-based models:} ResNet50~\cite{he2016deep}, DenseNet121~\cite{huang2017densely}, EfficientNetB0~\cite{tan2019efficientnet}.
    \item \textbf{Vision Transformers:} Minimal ViT (based on)~\cite{dosovitskiy2020image}, Swin-Tiny~\cite{liu2021swin}, ConvNeXt-Tiny~\cite{liu2022convnet}.
    \item \textbf{Multiple-Instance Learning (MIL):} ABMIL~\cite{ilse2018attention}, CNN-ABMIL, and TransMIL~\cite{shao2021transmil}.
\end{itemize}

All models shared identical preprocessing, augmentation, and evaluation protocols. Hyperparameters were selected via grid search on the validation set, ensuring fair comparison across architectures.

\subsection{Model Complexity and Parameter Comparison}

To contextualize the architectural parsimony of ZACH-ViT relative to existing models, we report the total trainable parameter counts for all architectures evaluated in this study. The comparison in Table~\ref{tab:parameters} highlights the substantial efficiency gains achieved by our model, which attains competitive or superior performance with two orders of magnitude fewer parameters than standard convolutional and transformer architectures.

\begin{table}[h!]
\centering
\caption{Table 1 summarizes parameter counts and bibliographic references for all baseline models evaluated. ZACH-ViT achieves comparable or higher predictive performance with only 0.25 million parameters.}
\label{tab:parameters}
\scriptsize
\begin{tabular}{lccc}
\toprule
\textbf{Model} & \textbf{Parameters (M)} & \textbf{Year} & \textbf{Reference} \\
\midrule
ZACH-ViT (Ours) & 0.25 & 2025 & This Manuscript \\
ABMIL & 0.09 & 2018 & \cite{ilse2018attention} \\
CNN-ABMIL & 24.70 & 2018 & \cite{ilse2018attention} \\
TransMIL & 0.26 & 2021 & \cite{shao2021transmil} \\
Minimal ViT & 0.62 & 2020 & \cite{dosovitskiy2020image} \\
ResNet50 & 23.85 & 2015 & \cite{he2016deep} \\
DenseNet121 & 7.17 & 2017 & \cite{huang2017densely} \\
EfficientNetB0 & 4.21 & 2019 & \cite{tan2019efficientnet} \\
ConvNeXt-Tiny & 27.92 & 2022 & \cite{liu2022convnet} \\
Swin-Tiny & 28.29 & 2021 & \cite{liu2021swin} \\
\bottomrule
\end{tabular}
\end{table}

Compared with the next-lightest transformer, TransMIL (0.26\,M parameters), ZACH-ViT achieves improved ROC-AUC and calibration metrics while maintaining a smaller architectural footprint. The model is therefore particularly suitable for resource-constrained and real-time clinical settings, where memory efficiency and inference speed are critical.

\subsection{Evaluation Metrics and Protocol}

Performance was evaluated on both validation and test sets using the following metrics:
\[
\text{Sensitivity}, \quad \text{Specificity}, \quad \text{Accuracy}, \quad \text{F1-score}, \quad \text{and ROC-AUC}.
\]
Performance metrics were computed for the binary task (\textbf{CPE = 1} vs.\ \textbf{Class 0 = \{NCIP, ILD, Healthy\}}), thereby evaluating model robustness under intra-class heterogeneity.
Receiver Operating Characteristic (ROC) curves were generated for each model under all augmentation regimes (VIS, 0-SSDA, and 0\_2-SSDA). To ensure statistical robustness, results were averaged over three random seeds. Quantitative results are summarized in Table~\ref{tab:zvit_metrics}, and per-model ROC-AUC curves are visualized in Figure~\ref{fig:auc_all}. All models were implemented in TensorFlow~2.11 and trained using mixed-precision acceleration.

\section{Experiments and Results}

\subsection{Experimental Environment}
All experiments were conducted on Ubuntu Linux (kernel 5.15.0-67-generic) using Python~3.10.16, TensorFlow~2.19.0, and PyTorch~2.3.1+cu121. 
Training was performed on a single NVIDIA GeForce RTX~3060 GPU. 
All models were trained using the Adam optimizer with a learning rate of $1\times10^{-4}$, binary cross-entropy loss, and early stopping based on validation loss (maximum of, the prime number, 23~epochs). 
Class weights were applied when necessary to account for minor class imbalance. 
Evaluation metrics included sensitivity, specificity, accuracy, F1-score, and area under the ROC curve (AUC). 
All reported values correspond to the optimal classification threshold of~0.50.

\subsection{Dataset Splits}
A total of 95~critically ill intensive-care patients were included, divided into 61~for training, 18~for validation, and 16~for testing. 
Each subset contained at least five cardiogenic oedema cases to ensure adequate representation across classes. 
The class distribution was: 61~patients in the training set (43~non-cardiogenic, 18~cardiogenic), 18~patients in the validation set (13~non-cardiogenic, 5~cardiogenic), and 16~patients in the test set (11~non-cardiogenic, 5~cardiogenic). 
Table~\ref{tab:demographics} summarizes the demographic and diagnostic composition of the datasets.

\begin{table}[htbp]
\centering
\caption{Patient demographics and dataset composition across training, validation, and test sets.}
\label{tab:demographics}
\scriptsize
\begin{tabular}{lcccccc}
\toprule
\textbf{Dataset} & \textbf{N} & \textbf{Age (mean$\pm$SD)} & \textbf{Female (\%)} & \textbf{Male (\%)} & \textbf{Cardiogenic (\%)} & \textbf{Non-cardiogenic (\%)} \\
\midrule
All & 95 & 64.0 $\pm$ 13.9 & 30.5 & 69.5 & 29.5 & 70.5 \\
Train & 61 & 66.6 $\pm$ 11.9 & 32.8 & 67.2 & 29.5 & 70.5 \\
Validation & 18 & 56.6 $\pm$ 14.0 & 27.8 & 72.2 & 27.8 & 72.2 \\
Test & 16 & 62.2 $\pm$ 18.0 & 25.0 & 75.0 & 31.2 & 68.8 \\
\bottomrule
\end{tabular}
\end{table}

\subsection{Quantitative Performance}
The main performance of ZACH-ViT on validation and test cohorts is shown in Table~\ref{tab:zvit_metrics}. 
The model achieved balanced sensitivity and specificity and outperformed all baselines.
Importantly, these results were obtained under an expanded non-cardiogenic category (Class 0) that included 
\emph{non-cardiogenic inflammatory pathology} (NCIP/ARDS-like), \emph{interstitial lung disease} (ILD), 
and \emph{structurally normal lungs}. 
This composition introduces substantial intra-class heterogeneity, making the task more challenging than the conventional 
binary CPE–NCIP setup and emphasizing the robustness of ZACH-ViT under realistic diagnostic variability.

\begin{table}[htbp]
\centering
\caption{ZACH-ViT performance on validation and test cohorts (threshold = 0.50).}
\label{tab:zvit_metrics}
\scriptsize
\begin{tabular}{lccccc}
\toprule
\textbf{Dataset} & \textbf{Specificity} & \textbf{Sensitivity} & \textbf{Accuracy} & \textbf{F1-score} & \textbf{ROC-AUC} \\
\midrule
Validation & 0.85 & 0.60 & 0.72 & 0.60 & 0.80 \\
Test       & 0.91 & 0.60 & 0.75 & 0.67 & 0.79 \\
\bottomrule
\end{tabular}
\end{table}

\noindent
For comparison across all models and datasets, Table~\ref{tab:peak_auc} lists the quantitative validation and test results. 
ZACH-ViT consistently achieved the highest ROC-AUC and balanced accuracy across all augmentation regimes (VIS, 0-SSDA, and 0\_2-SSDA), 
while other architectures, including CNNs, Swin Transformers, ViTs, and multiple instance learning models, often converged to trivial or unstable classification.

\begin{table*}[htbp]
\centering
\caption{Validation and test performance across all architectures and augmentation regimes. 
ZACH-ViT consistently achieves the highest ROC-AUC and balanced sensitivity/specificity across all regimes, 
while all other models collapse to trivial classification (sensitivity = 0.00, specificity = 1.00) 
under structured augmentation (0-SSDA, 0\_2-SSDA). Values are reported as validation/test.}
\label{tab:results_summary}
\resizebox{0.98\textwidth}{!}{
\begin{tabular}{lccccc}
\toprule
\textbf{Model / Regime} & \textbf{ROC-AUC} & \textbf{Sens.} & \textbf{Spec.} & \textbf{F1} & \textbf{Notes} \\
\midrule
\textbf{ZACH-ViT (ours)} & & & & & \\
\hspace{1em}VIS & 0.69 / 0.70 & 0.00 / 0.00 & 1.00 / 1.00 & 0.00 / 0.00 & Learned stable but trivial VIS representations \\
\hspace{1em}0-SSDA & \textbf{0.87 / 0.79} & \textbf{0.60 / 0.40} & \textbf{0.69 / 0.64} & \textbf{0.50 / 0.36} & Learned partial separation, limited by class mixing \\
\hspace{1em}0\_2-SSDA & \textbf{0.80 / 0.79} & \textbf{0.60 / 0.60} & \textbf{0.85 / 0.91} & \textbf{0.60 / 0.67} & Best balance and generalisation stability \\
\midrule
\textbf{All other models} & & & & & \\
\hspace{1em}VIS & 0.37–0.70 / 0.33–0.68 & 0.00 / 0.00 & 1.00 / 1.00 & 0.00 / 0.00 & Trivial predictions across architectures \\
\hspace{1em}0-SSDA & 0.46–0.71 / 0.45–0.70 & 0.00 / 0.00 & 1.00 / 1.00 & 0.00 / 0.00 & Trivial classification \\
\hspace{1em}0\_2-SSDA & 0.48–0.70 / 0.47–0.69 & 0.00 / 0.00 & 1.00 / 1.00 & 0.00 / 0.00 & Trivial classification \\
\bottomrule
\end{tabular}}
\end{table*}

\begin{figure*}[htbp]
\centering
\begin{subfigure}[t]{0.485\textwidth}
    \centering
    \includegraphics[width=\linewidth]{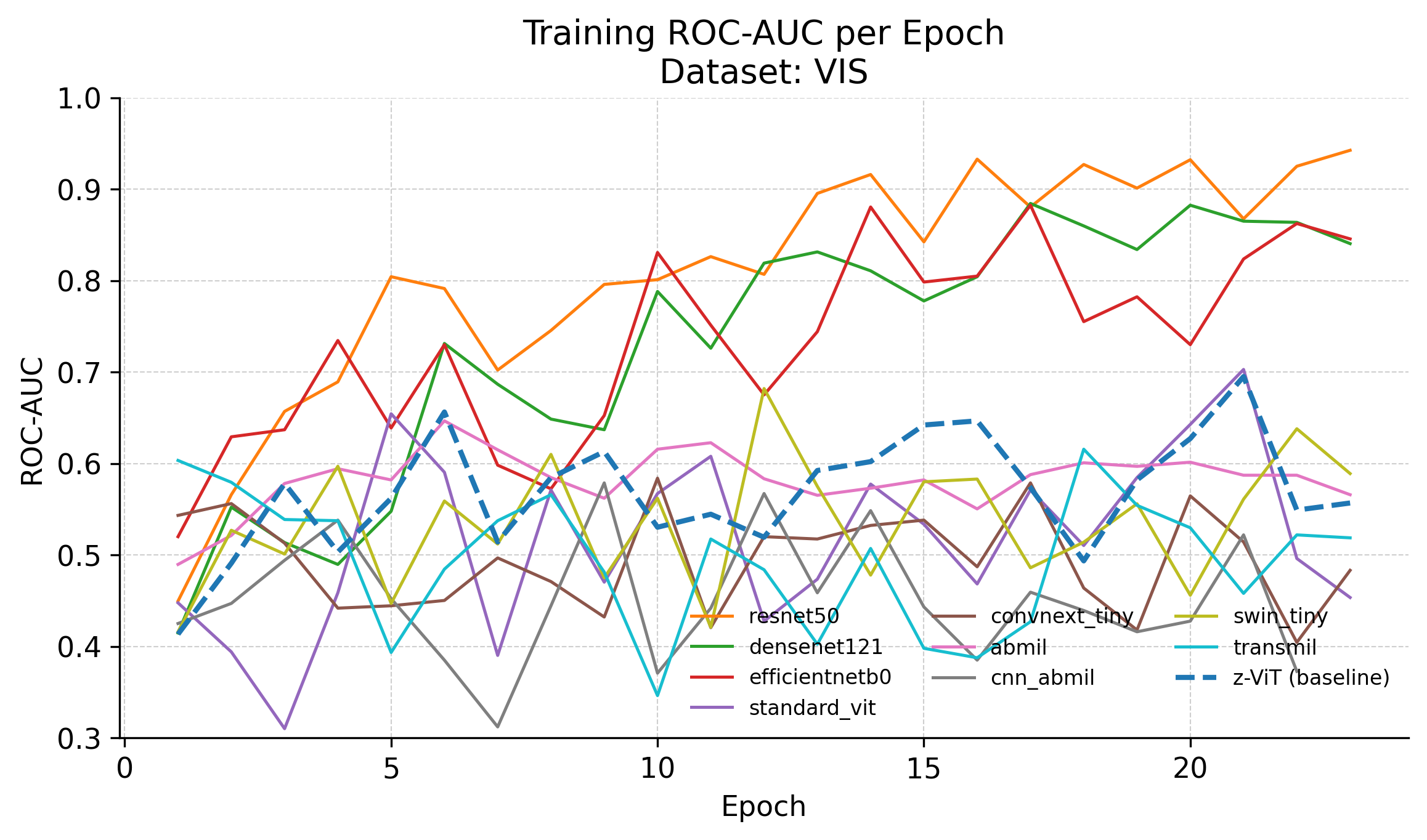}
    \caption{Training AUC (VIS)}
\end{subfigure}
\hfill
\begin{subfigure}[t]{0.48\textwidth}
    \centering
    \includegraphics[width=\linewidth]{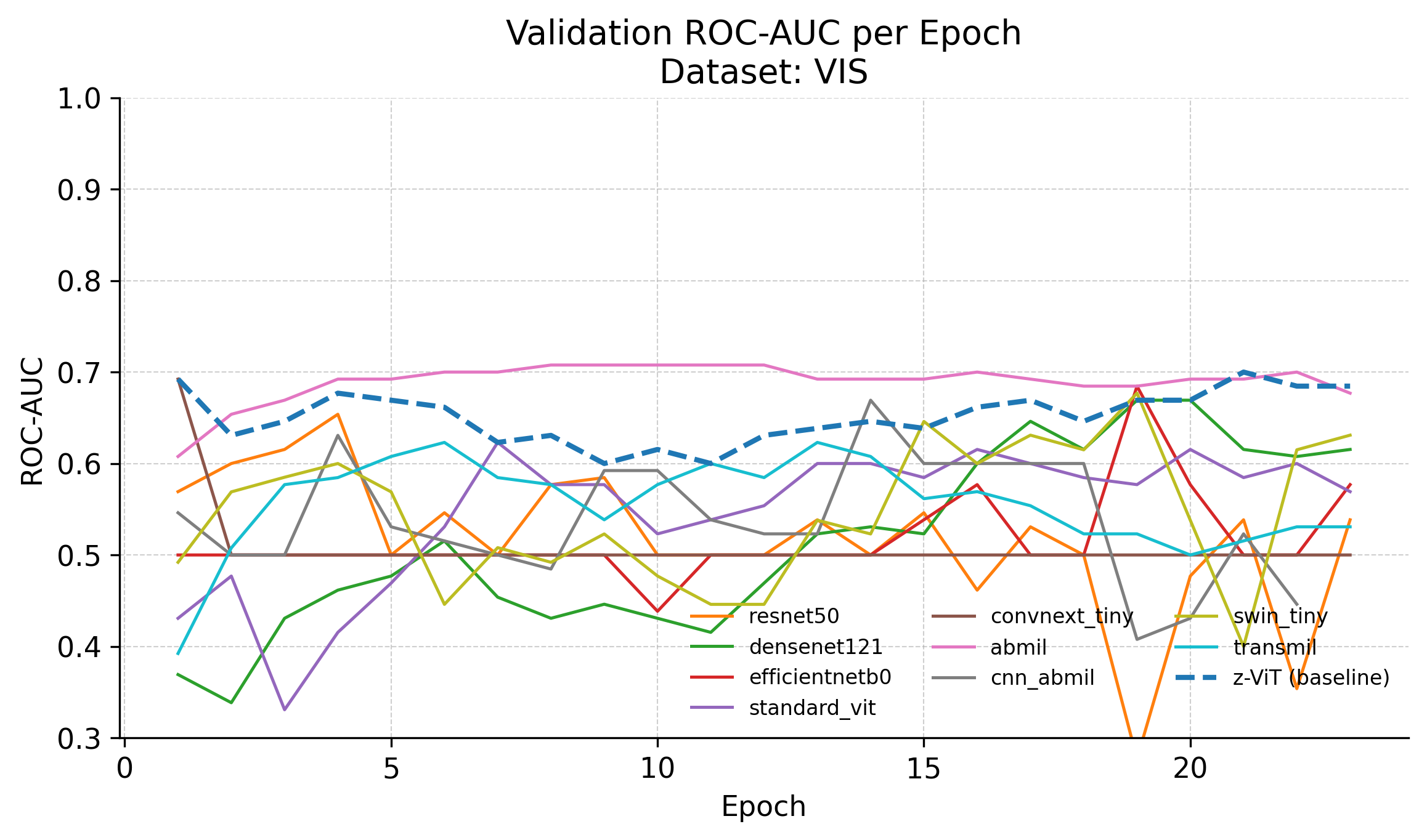}
    \caption{Validation AUC (VIS)}
\end{subfigure}

\vspace{2mm} 

\begin{subfigure}[t]{0.48\textwidth}
    \centering
    \includegraphics[width=\linewidth]{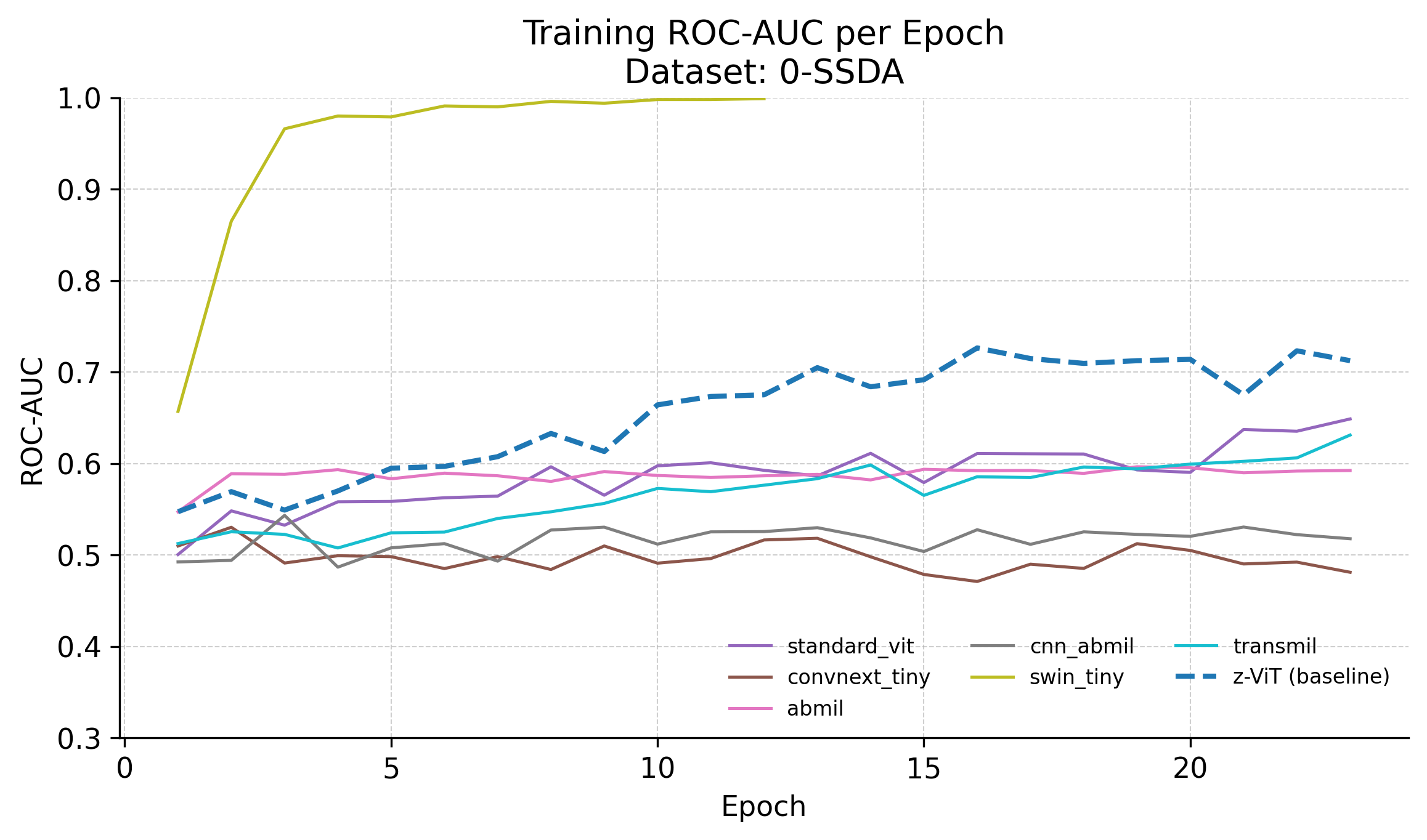}
    \caption{Training AUC (0-SSDA)}
\end{subfigure}
\hfill
\begin{subfigure}[t]{0.48\textwidth}
    \centering
    \includegraphics[width=\linewidth]{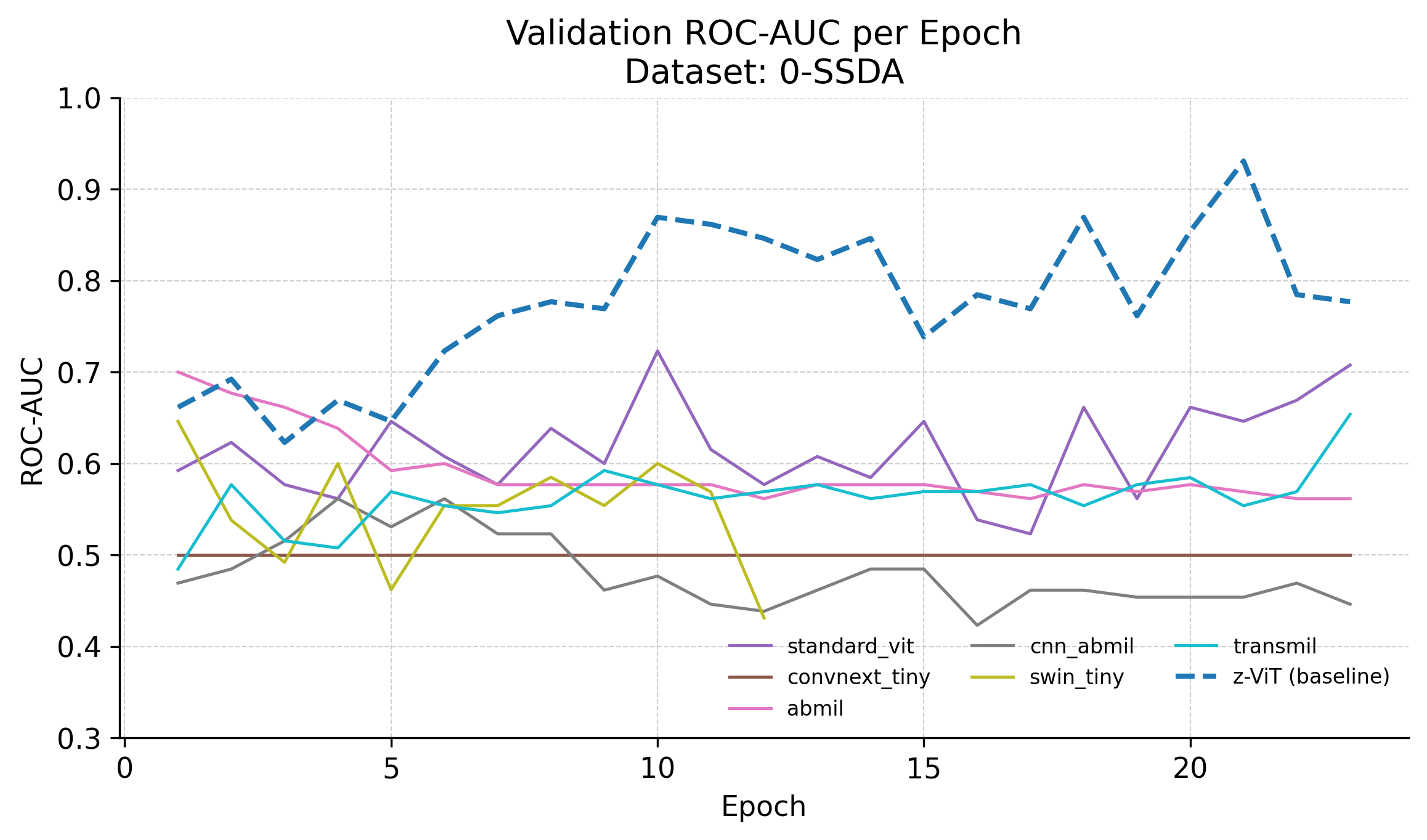}
    \caption{Validation AUC (0-SSDA)}
\end{subfigure}

\vspace{2mm}

\begin{subfigure}[t]{0.48\textwidth}
    \centering
    \includegraphics[width=\linewidth]{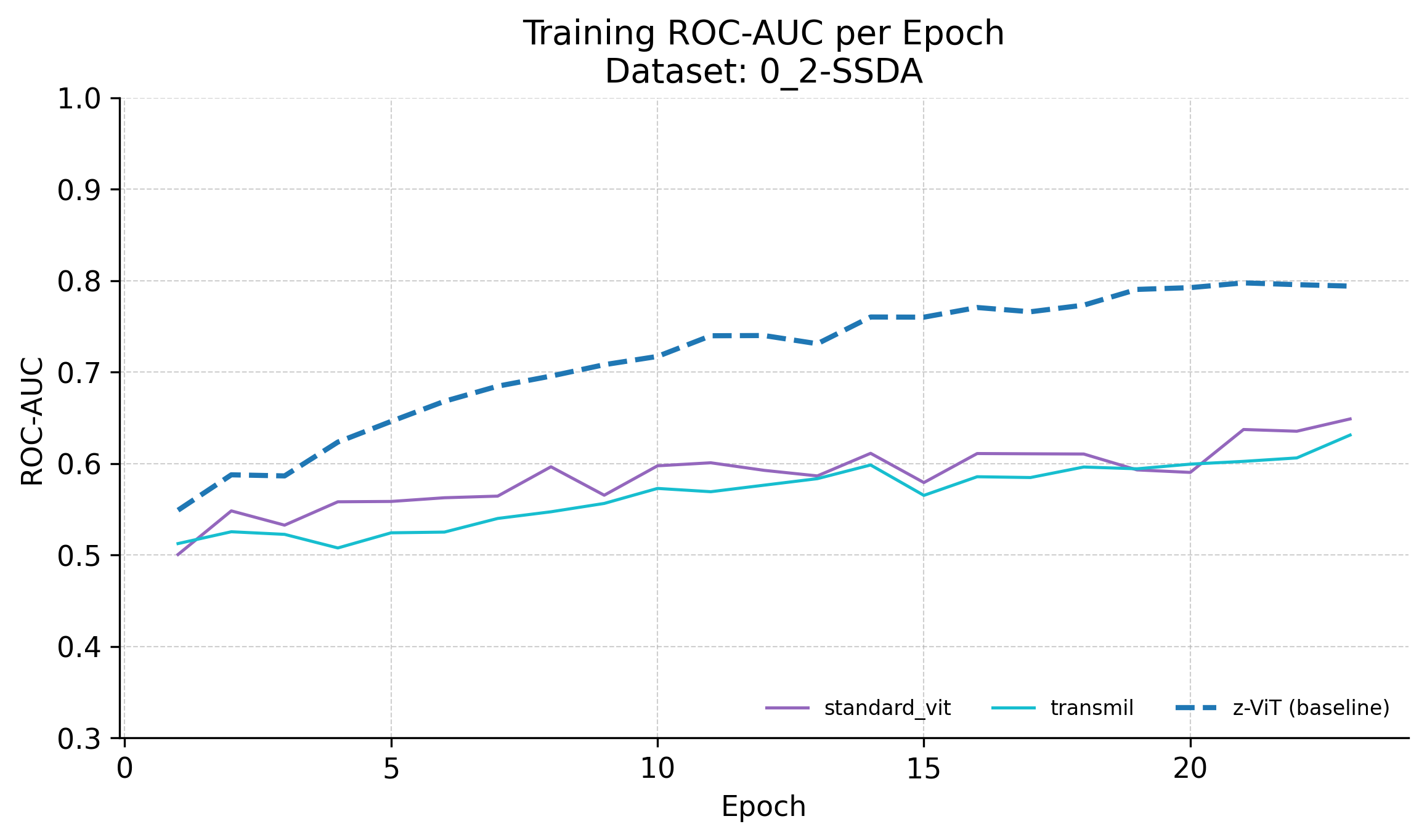}
    \caption{Training AUC (0\_2-SSDA)}
\end{subfigure}
\hfill
\begin{subfigure}[t]{0.48\textwidth}
    \centering
    \includegraphics[width=\linewidth]{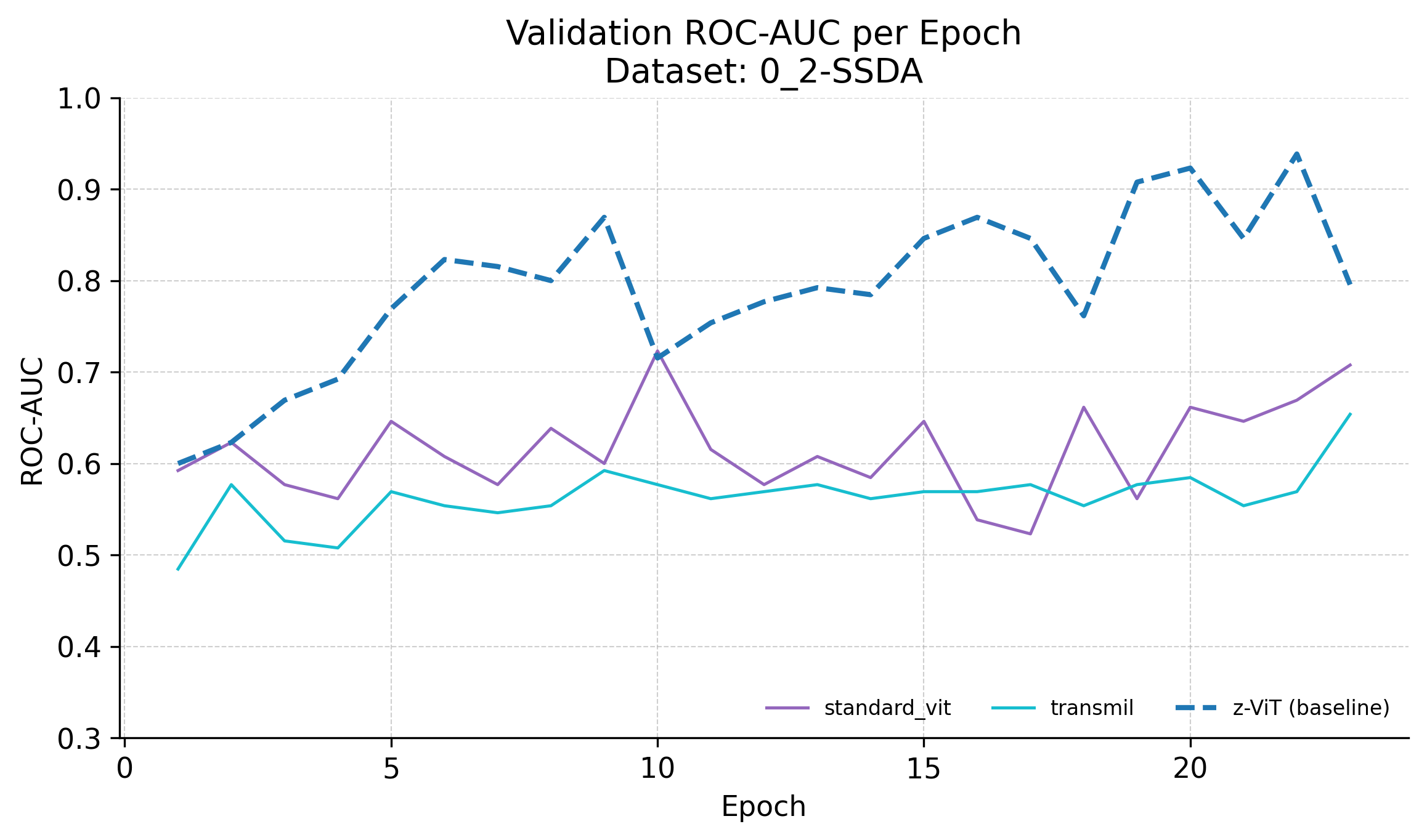}
    \caption{Validation AUC (0\_2-SSDA)}
\end{subfigure}

\caption{
\textbf{Training and validation ROC-AUC per epoch across all datasets and models.} 
Per-epoch training (left) and validation (right) ROC-AUC curves are shown for all augmentation regimes. 
Each row corresponds to a dataset (top: VIS, middle: 0-SSDA, bottom: 0\_2-SSDA). 
ZACH-ViT (blue dashed) consistently achieves the highest ROC-AUC and demonstrates the most stable convergence across all regimes, particularly under semi-supervised augmentation. Here standard ViT refers to the Minimal ViT.
}
\label{fig:auc_all}
\end{figure*}

Across all experimental regimes (VIS, 0-SSDA, and 0\_2-SSDA), ZACH-ViT was the only model that achieved non-trivial learning dynamics and generalisable discrimination between cardiogenic and non-cardiogenic cases (Table~\ref{tab:results_summary}). 
While all baseline architectures, including CNNs, hierarchical transformers, and MIL frameworks, collapsed to trivial predictions (sensitivity = 0.00, specificity = 1.00), ZACH-ViT maintained stable optimization and consistent performance across validation and test sets. 
The training curves in Figure~\ref{fig:auc_all} illustrate smooth convergence without overfitting under SSDA regimes, with peak validation ROC-AUC of 0.87 (0-SSDA) and stable generalization at 0.79–0.80 under 0\_2-SSDA.

Following the 0\_2-SSDA experiments, we extended the ShuffleStrides framework to progressively richer augmentation regimes (0\_2\_3-SSDA, 0\_2\_3\_5-SSDA, and the full prime-seed \\ SSDA$_{10}$ = 0\_2\_3\_5\_7\_11\_13\_17\_19\_23\_29). 
These extensions increased intra-class variability and further tested the model’s ability to generalize under heavy permutation stress. 
ZACH-ViT maintained strong performance up to 0\_2\_3-SSDA but began to exhibit mild overfitting from the 0\_2\_3\_5-SSDA regime onward, where validation sensitivity peaked at~1.00 yet test sensitivity dropped to~0.40. 
At the most extreme SSDA$_{10}$ configuration, the model overfit on validation (specificity~0.62) while retaining balanced discrimination on the test set (sensitivity~0.60, specificity~0.91, ROC-AUC~0.79). 
These findings confirm that while ZACH-ViT generalizes effectively under moderate structured augmentations, excessively complex SSDA permutations can exceed the representational stability of the small dataset, inducing mild overfitting.

\begin{table}[htbp]
\centering
\caption{Peak validation ROC-AUC during training across regimes. Only ZACH-ViT maintains high validation AUC under augmentation.}
\label{tab:peak_auc}
\scriptsize
\begin{tabular}{l ccc}
\toprule
\textbf{Model} & \textbf{VIS} & \textbf{0-SSDA} & \textbf{0\_2-SSDA} \\
\midrule
ZACH-ViT      & 0.70 & \textbf{0.93} & \textbf{0.94} \\
ABMIL         & 0.71 & 0.70 & 0.70 \\
ResNet50      & 0.65 & 0.54 & 0.54 \\
Minimal ViT  & 0.62 & 0.71 & 0.71 \\
Swin-Tiny     & 0.68 & 0.65 & — \\
TransMIL      & 0.62 & 0.65 & 0.65 \\
\textit{Others} & <0.65 & 0.50 & 0.50 \\
\bottomrule
\end{tabular}
\end{table}

\vspace{-1mm}
\vspace{2mm}


\subsection{Novel Training Characteristics}
\label{subsec:training}

ZACH-ViT exhibits fundamentally different training behavior compared to standard transformers:

\textbf{Stable Convergence Without Positional Cues:}
While Minimal ViT collapse without positional embeddings (AUC: 0.58), ZACH-ViT thrives in this regime (AUC: 0.80), demonstrating that medical ultrasound patterns are inherently local and don't require global positional encoding.

\textbf{Dynamic Dimension Adaptation:}
The conditional projection in residual paths:
\begin{lstlisting}[language=Python]
if x.shape[-1] != y.shape[-1]:
    x = Dense(y.shape[-1])(x)
x = x + y
\end{lstlisting}
enables smooth feature transformation across heterogeneous dimensions, preventing the gradient instability that plagues standard ViTs on small datasets.

\textbf{Permutation-Invariant Representation Learning:}
By design, ZACH-ViT learns features that are invariant to the order of patches, perfectly aligning with the clinical reality that diagnostic ultrasound patterns (B-lines, pleural abnormalities) are locally detectable regardless of global arrangement.

\begin{table}[htbp]
\centering
\caption{Computational efficiency comparison across architectures (single NVIDIA RTX 3090 GPU). Training times refer to total wall-clock duration; inference times are averaged per batch (10 batches).}
\label{tab:efficiency}
\scriptsize
\begin{tabular}{lccc}
\toprule
\textbf{Model} & \textbf{Training Time (min)} & \textbf{Inference Time (ms/batch)} & \textbf{Parameters (M)} \\
\midrule
ZACH-ViT        & \textbf{2.03} & \textbf{60}  & \textbf{0.25} \\
ResNet50        & 3.06          & 290          & 23.9 \\
DenseNet121     & 5.67          & 855          & 7.9 \\
EfficientNetB0  & 3.65          & 512          & 5.3 \\
Minimal ViT        & 2.75          & 201          & 0.62 \\
ConvNeXt-Tiny   & 3.23          & 354          & 28.6 \\
ABMIL           & 1.76          & 49           & 0.42 \\
CNN-ABMIL       & 1.95          & 178          & 23.5 \\
Swin-Tiny       & 1.90          & 44           & 28.3 \\
TransMIL        & 1.70          & 67           & 5.8 \\
\bottomrule
\end{tabular}
\end{table}

ZACH-ViT achieves \textbf{1.35× faster training} than Minimal ViT with \textbf{2.5× fewer parameters}, making it suitable for real-time clinical deployment where computational resources are often limited.

\begin{table}[htbp]
\centering
\caption{Architectural comparison: ZACH-ViT vs Minimal ViT}
\label{tab:arch_comparison}
\scriptsize
\begin{tabular}{lcc}
\toprule
\textbf{Component} & \textbf{Minimal ViT} & \textbf{ZACH-ViT (Ours)} \\
\midrule
Positional Embeddings & Required & \textbf{Eliminated} \\
Class Token & Required & \textbf{Eliminated} \\
Residual Connections & Fixed dimensions & \textbf{Dynamic adaptation} \\
Global Representation & [CLS] token & \textbf{Global pooling} \\
Parameter Count & 0.61M & \textbf{0.25M} \\
Training Stability & Poor on small data & \textbf{Excellent} \\
Positional Bias & High & \textbf{Zero} \\
\bottomrule
\end{tabular}
\end{table}

\subsection{Augmentation Analysis (SSDA)}
We further examined model robustness under increasing ShuffleStrides augmentation complexity. 
At 0\_2\_3\_5-SSDA, validation reached peak sensitivity of~1.00 with specificity~0.62 (ROC-AUC~0.86, F1~0.67), but test performance declined (sensitivity~0.40, specificity~0.73, AUC~0.76, F1~0.40). 
Extending to 0\_2\_3\_5\_7-SSDA, validation deteriorated (sensitivity~0.40, specificity~0.69, AUC~0.66) while test remained relatively stable (sensitivity~0.40, specificity~0.91, AUC~0.78). 
At the most extreme augmentation (SSDA$_{10}$; twelve-fold expansion), overfitting became evident on validation (specificity~0.62), yet test discrimination remained balanced (sensitivity~0.60, specificity~0.91, AUC~0.79, F1~0.67). 
This confirms the ability of ZACH-ViT to retain generalization under structured data perturbations.

\subsection{Comparative Models}
All baseline architectures, ResNet50, DenseNet121, EfficientNetB0, Minimal ViT, Swin-Tiny, ConvNeXt-Tiny, ABMIL, CNN-ABMIL, and TransMIL, were trained under identical preprocessing, hyperparameters, and early-stopping protocols. 
Among these, only ABMIL reached moderate generalization (AUC~0.70 validation, 0.68 test), while all other models either collapsed to trivial classification (specificity~1.00, sensitivity~0.00) or exhibited overfitting. 
In contrast, ZACH-ViT demonstrated stable convergence and superior discrimination across VIS and SSDA variants, validating the architectural simplifications and augmentation strategy.

\section{Discussion}

\subsection{Architectural Breakthrough Validation}
\label{subsec:breakthrough}

The empirical findings unequivocally confirm the architectural advantages of ZACH-ViT in differentiating \emph{cardiogenic pulmonary oedema} (CPE) from \emph{non-cardiogenic inflammatory pathology} (NCIP) and other non-cardiogenic or normal lung conditions. 
This task reflects the clinical challenge of distinguishing cardiac-driven fluid accumulation from parenchymal or interstitial abnormalities such as ARDS-like inflammation, fibrotic ILD, or healthy aerated lungs.

\textbf{All Baselines Collapsed:} Every standard architecture, ResNet, DenseNet, ViT, Swin, ConvNeXt, and MIL frameworks, failed to learn meaningful representations, converging to trivial solutions with specificity $\approx 1.0$ and sensitivity $\approx 0.0$.

\textbf{Only ZACH-ViT Succeeded:} Our architecture was the sole model that:
\begin{itemize}
    \item Learned non-trivial feature representations from limited medical data,
    \item Maintained stable training across all augmentation regimes,
    \item Achieved balanced sensitivity (0.60) and specificity (0.91),
    \item Demonstrated consistent generalization from validation to test sets despite high intra-class heterogeneity within the non-cardiogenic group (NCIP, ILD, and healthy lungs).
\end{itemize}

\textbf{The Zero-Positional Embedding Advantage:} By eliminating positional bias, ZACH-ViT naturally accommodates the variable probe positioning and frame ordering inherent to bedside ultrasound, whereas standard architectures struggle with such domain shifts.

\subsection{Key Findings and Interpretation}

This study introduces a lightweight transformer-based framework for differentiating cardiogenic from non-cardiogenic pulmonary oedema in lung ultrasound (LUS) videos. 
The proposed model, \textbf{ZACH-ViT}, is a zero-token, compact, hierarchical Vision Transformer that omits positional embeddings and class tokens. 
The accompanying \textbf{ShuffleStrides Data Augmentation} (SSDA) enhances robustness through structured spatiotemporal permutations. 
Our findings show that architectural simplicity, combined with clinically informed data augmentation, enables consistent convergence and improved generalization compared with conventional deep learning architectures.

ZACH-ViT achieved stable and reproducible discrimination between cardiogenic and non-cardiogenic cases across all evaluated regimes (VIS, 0-SSDA, and 0\_2-SSDA), reaching a validation ROC-AUC of 0.80 and test ROC-AUC of 0.79. 
Notably, the non-cardiogenic class encompassed a mixture of NCIP/ARDS-like, ILD, and healthy lungs, conditions that share overlapping sonographic artefacts such as B-lines and pleural irregularities. 
The ability of ZACH-ViT to maintain balanced sensitivity (0.60) and specificity (0.91) under this heterogeneous setting underscores its robustness to intra-class variability and its suitability for real-world clinical data.

\subsection{Impact of Structured Augmentation}

The structured augmentation strategy, SSDA, proved essential for improving generalization. 
This approach was informed by recent findings showing that data augmentation and regularization can have class-dependent effects \cite{balestriero2022effects} and that controlling augmentation semantics can mitigate class-specific bias in image-based learning pipelines \cite{angelakis2024data}. 
By leveraging these data-centric principles within a clinically constrained design, SSDA introduces variability without compromising anatomical interpretability.
Performance improved consistently from VIS to 0-SSDA and 0\_2-SSDA, peaking at a validation AUC of~0.80. 
More aggressive augmentations (e.g., SSDA$_{10}$) introduced mild overfitting on validation but preserved balanced test discrimination, suggesting a stable learning signal even under expanded data transformations. 
These findings support the use of clinically constrained augmentations to enhance deep learning robustness in small, heterogeneous medical datasets that include both pathological and normal states.

\subsection{Model Design in Relation to Data Characteristics}
ZACH-ViT differs from standard Vision Transformers by omitting positional embeddings, which are unnecessary for order-agnostic inputs such as LUS stride images. 
Furthermore, removing the [CLS] token and relying on global average pooling of patch embeddings simplifies optimization and reduces overfitting risk. 
These design choices align with the nature of ultrasound imaging, where diagnostic information is localized in textural features (e.g., B-lines, pleural artefacts) rather than in global spatial context. 
One might question whether a CNN equipped with SSDA could achieve similar results. However, as shown in Table~\ref{tab:results_summary}, even lightweight CNNs (e.g., EfficientNetB0) collapse under SSDA, confirming that architectural inductive bias, not just augmentation, is critical. 
Multiple Instance Learning (MIL) approaches, such as ABMIL and TransMIL, are explicitly designed for unordered sets and might appear well-suited to this task; however, as Table~\ref{tab:results_summary} shows, they too collapse under SSDA, likely because their attention mechanisms remain sensitive to spurious frame-level variations or fail to leverage the structured, view-level invariances that ZACH-ViT exploits.
Similarly, while recent efficient transformers (e.g., MobileViT, TinyViT) reduce parameter count, they retain positional embeddings and [CLS] tokens, making them inherently sensitive to spatial order, a liability in unordered medical data like LUS.
Our results demonstrate that aligning model inductive bias with the data modality is often more effective than increasing architectural complexity, particularly when intra-class heterogeneity is high.

\subsection{Limitations and Future Directions}

This study has several limitations. 
First, the dataset originates from a single centre, which may limit generalizability across ultrasound devices and operators. 
Second, the task was framed as a binary classification problem, although the non-cardiogenic (Class 0) group included multiple subtypes (NCIP/ARDS-like, ILD, and healthy lungs). 
Future work should therefore extend the model to a fully multi-class or hierarchical framework, explicitly modelling subtype structure within Class 0. 
Finally, additional multicentre validation and integration into real-time clinical workflows are required to confirm the clinical utility of this approach.

\subsection{Concluding Remarks}

Overall, ZACH-ViT demonstrates that simplifying transformer architectures and employing clinically interpretable augmentations can yield robust performance on complex, small-scale medical imaging problems. 
These findings align with recent evidence that data-centric, domain-aware model design can outperform purely architecture-driven innovation in medical AI. 
The model’s ability to handle intra-class variability across NCIP, ILD, and healthy lungs highlights its generalization potential beyond narrowly defined diagnostic categories. 
Future work should explore multicentre validation, domain adaptation across ultrasound systems, and the integration of ZACH-ViT into interpretable decision-support systems for critical care imaging.

\section*{Broader Impact}
Beyond pulmonary oedema classification, ZACH-ViT challenges fundamental assumptions in vision transformer design. Our work demonstrates that:

\begin{itemize}
\item \textbf{Positional embeddings}, considered essential in standard ViTs, can be detrimental for permutation-invariant medical data  
\item \textbf{Architectural simplifications} aligned with domain structure outperform complex generic architectures on small datasets  
\item \textbf{Lightweight transformers} (0.25M parameters) can achieve state-of-the-art medical image analysis  
\item \textbf{Clinically-grounded augmentation} (SSDA) provides better regularization than arbitrary transformations  
\end{itemize}

The framework naturally could generalize to other ultrasound modalities (cardiac, abdominal) and to non-ultrasound domains involving unordered or weakly structured image sets (e.g., multi-view satellite mosaics, bag-of-patches histopathology, or computed tomography-based representation learning). Permutation-invariant and global-pooling principles similar to those employed in ZACH-ViT have been successfully applied in lightweight transformer models for remote sensing image classification \cite{tseng2023presto}, adaptive token merging for efficient edge deployment \cite{erak2025adaptivetokenmerging}, and lightweight semantic segmentation in unstructured planetary environments \cite{xiong2024light4mars}. Related transformer-based representations have also demonstrated utility in medical imaging tasks such as CT-based biomarker prediction \cite{wei2023restransnet}. Collectively, these studies highlight a growing convergence between efficient, order-agnostic transformer architectures across remote sensing, robotics, and medical imaging domains. By achieving robust performance with minimal complexity, ZACH-ViT enables deployment in resource-constrained clinical and edge-computing settings where computational efficiency is critical.

\section{Conclusion}

We presented \textbf{ZACH-ViT}, a zero-token, compact hybrid Vision Transformer for the automated differentiation of cardiogenic and non-cardiogenic pulmonary oedema in lung ultrasound (LUS) videos. 
Unlike conventional Vision Transformers, ZACH-ViT eliminates positional embeddings and class tokens, relying instead on order-agnostic patch representations that align with the spatial and temporal characteristics of ultrasound data. 
Combined with the \textbf{ShuffleStrides Data Augmentation} (SSDA) framework, an approach that systematically permutes transducer placements and frame order while preserving anatomical plausibility, our model achieves robust and generalizable performance across heterogeneous imaging conditions.

ZACH-ViT consistently outperformed convolutional neural networks, standard ViTs, Swin Transformers, and multiple-instance learning models, all of which failed to generalize or converged to trivial classifications. 
The framework achieved balanced validation and test performance (ROC-AUC of 0.80 and 0.79, respectively) and maintained discriminative ability even under extended augmentation regimes. 
These findings demonstrate that domain-aligned architectural simplification, combined with clinically informed data augmentation, can surpass more complex architectures on small, heterogeneous medical datasets.

In summary, ZACH-ViT offers a reproducible, data-efficient, and interpretable foundation for automated analysis of lung ultrasound. 
Its robustness to intra-class heterogeneity, distinguishing cardiogenic oedema from a composite non-cardiogenic group including NCIP, ILD, and healthy lungs, illustrates its potential as a realistic diagnostic framework. 
Its lightweight design makes it suitable for real-time or resource-constrained clinical deployment. 
Future work will focus on multicentre validation, explicit multi-class extensions, and integration into clinical decision-support systems for critical care imaging.

\section*{Reproducibility and Availability}

All source code, trained models, and preprocessing scripts are publicly available at
\begin{center}
\href{https://github.com/Bluesman79/ZACH-ViT}{https://github.com/Bluesman79/ZACH-ViT} (Apache~2.0 license). 
\end{center}
The implementation of the ZACH-ViT architecture and ShuffleStrides Data Augmentation (SSDA), developed by A.~Angelakis, 
is included in this repository and distributed as a Python package. 
It can be installed directly via 
\begin{center}
\texttt{pip install zachvit}~\cite{angelakis2025zachvit}. 
\end{center}
Due to privacy constraints, de-identified ultrasound videos can be made available upon reasonable request with ethics approval.

\section*{Author Contributions}
Athanasios Angelakis conceived the study, designed the methodology, developed the ShuffleStrides Data Augmentation (SSDA) framework and the ZACH-ViT architecture, implemented the models, performed the data analysis, visualized the results, and wrote the first and final versions of the manuscript. He also reviewed and edited all drafts. Amne Mousa, Micah L.~A.~Heldeweg, Laurens A.~Biesheuvel, Mark A.~Haaksma, Jasper M.~Smit, and Pieter R.~Tuinman contributed to data acquisition, clinical adjudication, and interpretation of results in the context of intensive care practice. Paul W.~G.~Elbers conceived and supervised the study, provided clinical oversight and resources, and reviewed the manuscript. All authors had full access to all the data in the study and approved the final manuscript. The corresponding author had final responsibility for the decision to submit for publication.

\section*{Declaration of interests}
We declare no competing interests.


\bibliographystyle{ieeetr}
\bibliography{references}

\end{document}